\documentclass{article}

\usepackage[preprint, nonatbib]{neurips_2024}

\usepackage{bm}
\usepackage{caption}
\usepackage[utf8]{inputenc} 
\usepackage[T1]{fontenc}    
\usepackage{hyperref}       
\usepackage{url}            
\usepackage{booktabs}       
\usepackage{amsfonts}       
\usepackage{nicefrac}       
\usepackage{microtype}      
\usepackage{xcolor}         
\usepackage{graphicx}
\usepackage{amsmath}
\usepackage{colortbl}
\usepackage{makecell}
\usepackage{multirow}
\usepackage{amssymb}
\title{\textit{ContextGS}: Compact 3D Gaussian Splatting with Anchor Level Context Model}

\author{Yufei Wang$^1$ \quad Zhihao Li$^1$ \quad Lanqing Guo$^1$ \quad Wenhan Yang$^2$ \quad Alex C. Kot$^1$ \quad Bihan Wen$^1$ \\
$^1$ Nanyang Technological University, Singapore \\
$^2$ PengCheng Laboratory, China \\
\{yufei001, zhihao.li, lanqing.guo, eackot, bihan.wen\}@ntu.edu.sg \quad  yangwh@pcl.ac.cn
\\
\\
Homepage: \url{https://github.com/wyf0912/ContextGS}
}

\begin{document}

\def\eg{\textit{e.g.}}
\def\ie{\textit{i.e.}}

\def\f{\mathbf{f}}
\def\l{\mathbf{l}}
\def\v{\mathbf{v}}
\def\x{\mathbf{x}}
\def\s{\mathbf{s}}
\def\r{\mathbf{r}}
\def\c{\mathbf{c}}
\def\R{\mathbb{R}}
\def\O{\mathbf{O}}
\def\V{\mathbf{V}}
\def\z{\mathbf{z}}
\def\etal{\textit{et al.}}
\def\dvc{\vec{\mathbf{d}}_{c}}
\def\L{\mathcal{L}}

\newcommand{\wh}[1]{{\color{black} #1}}
\maketitle

\begin{figure}[h]
    \centering
    \vspace{-0.2cm}
    \includegraphics[width=\linewidth]{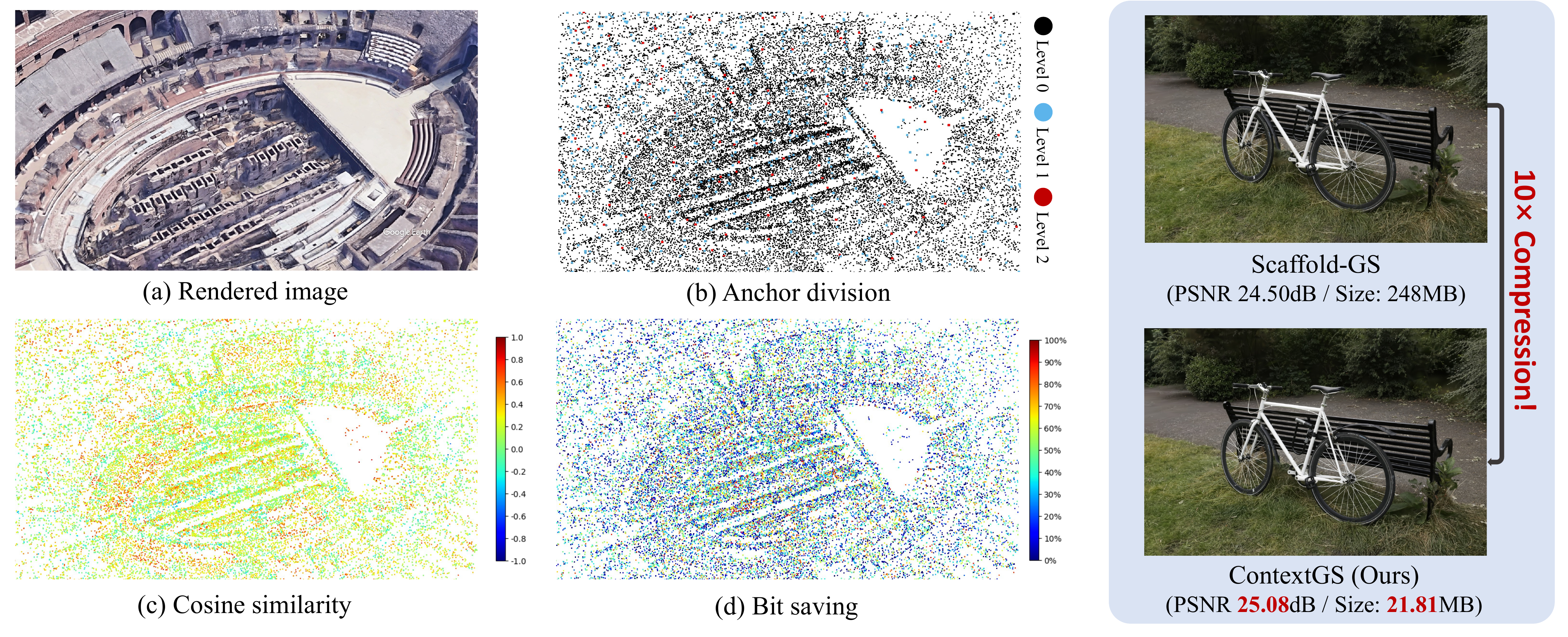}
    \caption{An illustration of the necessity of using autoregressive model in the anchor level. While Scaffold-GS~\cite{scaffold} greatly reduces the spatial redundancy among adjacent 3D neural Gaussians by grouping them and introducing a new data structure \textit{anchor} to capture their common features, spatial redundancy still exists among anchors. Our method, \textbf{ContextGS}, first proposes to reduce the spatial redundancy among anchors using an autoregressive model. We divide anchors into levels as shown in Fig.~(b) and the anchors from coarser levels are used to predict anchors in finer levels, \ie, \textcolor{red}{$\bullet$} predicts \textcolor{blue}{$\bullet$} then \textcolor{red}{$\bullet$}\textcolor{blue}{$\bullet$} together predict $\bullet$. Fig.~(c) verifies the spatial redundancy by calculating the cosine similarity between anchors in level $0$ and their context anchors in levels $1$ and $2$. Fig.~(d) displays the bit savings using the proposed anchor-level context model evaluated on our entropy coding based strong baseline built on Scaffold-GS~\cite{scaffold}. Compared with Scaffold-GS, we achieve better rendering qualities, faster rendering speed, and great size reduction of up to $15$ times averaged over all datasets we used.} 
    \label{fig:bit_saving}
\end{figure}

\begin{abstract}
Recently, 3D Gaussian Splatting (3DGS) has become a promising framework for novel view synthesis, offering fast rendering speeds and high fidelity. However, the large number of Gaussians and their associated attributes require effective compression techniques. 
Existing methods primarily compress neural Gaussians individually and independently, \ie, coding all the neural Gaussians at the same time, with little design for their interactions and spatial dependence. Inspired by the effectiveness of the context model in image compression, we propose the first autoregressive model at the anchor level for 3DGS compression in this work. We divide anchors into different levels and the anchors that are not coded yet can be predicted based on the already coded ones in all the coarser levels, leading to more accurate modeling and higher coding efficiency. To further improve the efficiency of entropy coding, \eg, to code the coarsest level with no already coded anchors, we propose to introduce a low-dimensional quantized feature as the hyperprior for each anchor, which can be effectively compressed. 
Our work pioneers the context model in the anchor level for 3DGS representation, yielding an impressive size reduction of over 100 times compared to vanilla 3DGS and 15 times compared to the most recent state-of-the-art work Scaffold-GS, while achieving comparable or even higher rendering quality.
\end{abstract}

\section{Introduction}

Over the past few years, novel view synthetic has rapidly progressed. As a representative work, Neural Radiance Field (NeRF)~\cite{NeRF} uses a Multilayer Perceptron (MLP) to predict the attributes of quired points in the 3D scene. While good rendering qualities are achieved, the dense querying process results in slow rendering, which greatly hinders their applications in practical scenarios. Significant efforts have been made to enhance training and rendering speeds, achieving notable progress through various techniques, such as factorization~\cite{TensoRF, K-planes, gao2023strivec, han2023multiscale} and hash grids~\cite{INGP, SHACIRA}. However, they still face challenges in the real-time rendering of large-scale scenes due to the intrinsic limitations of volumetric sampling.  
Recently, 3D Gaussian Splatting (3DGS)~\cite{3DGS} has achieved state-of-the-art (SOTA) rendering quality and speed. As an emerging alternative strategy for representing 3D scenes, 3DGS represents a 3D scene using a set of neural Gaussians initiated from Structure-from-Motion (SfM) with learnable attributes such as color, shape, and opacity. 
The 2D images can be effectively rendered using differentiable rasterization and end-to-end training is enabled. Meanwhile, benefiting from efficient CUDA implementation, real-time rendering is achieved.

Despite its success, 3DGS still encounters limitations in storage efficiency. Representing large scenes requires millions of neural Gaussian points, which demand several GBs of storage, \eg, an average of 1.6 GB for each scene in the BungeeNerf~\cite{BungeeNeRF} dataset. The huge storage overhead greatly hinders the applications of 3DGS~\cite{3DGS}, thus an efficient compression technique is required. However, the unorganized and sparse nature of these neural Gaussians makes it highly challenging to effectively reduce data redundancy. To address this issue, various techniques have been proposed to enhance the storage efficiency of 3D Gaussian models. For example, \cite{zhiwen, Joo, KLNavaneet, Simon} proposed to discretize continuous attributes of neural Gaussians to a cluster of attributes stored in the codebooks; \cite{zhiwen, Joo}~proposed to prune neural Gaussians with little effect. Entropy coding is also used to reduce the storage overhead by further encoding neural Gaussian features into bitstream~\cite{Sharath, chen2024hac, Joo, Wieland}. Although space utilization has greatly improved, they focus more on individually compressing each Gaussian point and do not well explore the relationship and reduce the spatial redundancy among neural Gaussians. 
To further reduce the spatial redundancy, most recently, \cite{scaffold} proposed to divide anchors into voxels and introduced an anchor feature for each voxel to grasp the common attributes of neural Gaussians in the voxel, \ie, the neural Gaussians are predicted by the anchor features. While the spatial dependency has been significantly reduced, as shown in Fig.~\ref{fig:bit_saving} (c), the similarity among anchors remains high in certain areas, indicating that spatial redundancy still exists.

To further enhance the coding efficiency of 3D scenes, we propose a novel framework named \textit{ContextGS} for 3DGS compression. Inspired by the effectiveness of context models~\cite{van2016conditional} in image compression~\cite{mentzer2018conditional}, we introduce an autoregressive model at the anchor level into 3DGS. Specifically, building on Scaffold-GS~\cite{scaffold}, we divide anchors into hierarchical levels and encode them progressively. Coarser level anchors are encoded first, and their decoded values are used to predict the distribution of nearby anchors at finer levels. This approach leverages spatial dependencies among adjacent anchors, allowing already decoded anchors to better predict the distribution of subsequent anchors, leading to significant improvements in coding efficiency. Additionally, anchors decoded at coarser levels can be directly used in the final fine-grained level, reducing storage overhead. To further enhance coding efficiency, especially for encoding the coarsest level anchors without already decoded ones, we employ a quantized hyperprior feature as an additional prior for each anchor.
Our contributions can be summarized as follows:
\begin{itemize}
    \item We propose the first context model for 3DGS \wh{at} the anchor level. By predicting the properties of anchors that are not coded yet given already coded ones, we greatly eliminate the spatial redundancy among channels. 
    \item We propose a unified compressing framework with the factorized prior, enabling end-to-end entropy coding of anchor features. Besides, a strategy for anchor layering is proposed, which allows already decoded anchors to quickly locate adjacent anchors that are to be decoded. Meanwhile, the proposed method avoids redundant coding of anchors by the proposed anchor forward.
    \item The experimental results on real-world datasets demonstrate the effectiveness of the proposed method compared with SOTA and concurrent works. On average across all datasets, our model achieves a compression ratio of 15$\times$ compared to the Scaffold-GS model we used as the backbone and 100$\times$ compared to the standard 3DGS model, while maintaining comparable or even enhanced fidelity.
\end{itemize}

\section{Related works}
\subsection{Neural radiance field and 3D Gaussian splatting}
Early 3D scene modeling often employs the Neural Radiance Field (NeRF)~\cite{NeRF} as a global approximator for 3D scene appearance and geometry. These approaches~\cite{barron2021mip, MaskDWT, takikawa2022variable} use a multi-layer perceptron (MLP) to implicitly represent the 3D scene by predicting attributes of queried points. However, the dense querying process results in extremely slow rendering. Various methods have been developed to speed up the rendering process significantly~\cite{deng2022depth, Plenoxels, reiser2021kilonerf}, such as plane factorization-based techniques like K-Planes~\cite{K-planes} and TensoRF~\cite{TensoRF}, and the use of hash grid features in InstantNGP~\cite{INGP}. While these methods enable high-quality rendering with a much smaller MLP compared to the vanilla NeRF, rendering a single pixel still requires numerous queries. This can lead to increased storage requirements for the grid-based features and difficulties in efficiently rendering empty space or large-scale scenes.
To achieve real-time and efficient rendering while maintaining high fidelity, 3DGS~\cite{3DGS} introduces an innovative approach by representing the scene explicitly with numerous learnable 3D Gaussians. By employing differentiable splatting and tile-based rasterization~\cite{raster}, 3DGS~\cite{3DGS} optimizes these Gaussians during training in an end-to-end manner. 

\subsection{Deep compression}
Despite the effectiveness of 3DGS~\cite{3DGS} in rendering speed and fidelity, the large number of Gaussians and their associated attributes result in significant storage overhead. Many techniques are proposed to reduce the storage requirements of 3DGS. For example, \cite{zhiwen, Joo} proposes to prune neural Guassians with minimal impact. \cite{zhiwen, Joo, KLNavaneet, Simon} propose to utilize codebooks to cluster Gaussian parameters. Entropy coding is also used in~\cite{Sharath, chen2024hac, Joo, Wieland} to encode neural Gaussians into bit streams by modeling their distributions. While remarkable performances are achieved, they mainly focus on improving the efficiency of a single neural Gaussian and neglect the spatial redundancy among neighbor neural Gaussians. Most recently, Scaffold-GS~\cite{scaffold} proposes to introduce an anchor level to capture common features of nearby neural Gaussians in the same voxel, and successive work~\cite{chen2024hac} demonstrates its effectiveness by further introducing hash-feature as a prior for entropy coding. However, \cite{chen2024hac} codes all the anchors at the same time, and its spatial redundancy can be further reduced. In the image compression task, an important category of methods to improve the coding efficiency is the context model~\cite{mentzer2018conditional, van2016conditional}, which greatly reduces the spatial redundancy by predicting the distribution of latent pixels based on an already coded one. Inspired by the context models used in compression, we propose to encode the anchor features in an autoregressive way, \ie, predict the anchor points from already coded ones at coarser levels. As far as we know, we are the first to reduce the storage redundancy of 3DGS using a context model at the anchor level. 

\section{Preliminary}
\begin{figure}
    \centering
    \includegraphics[width=\linewidth, clip, trim=0 20 0 0]{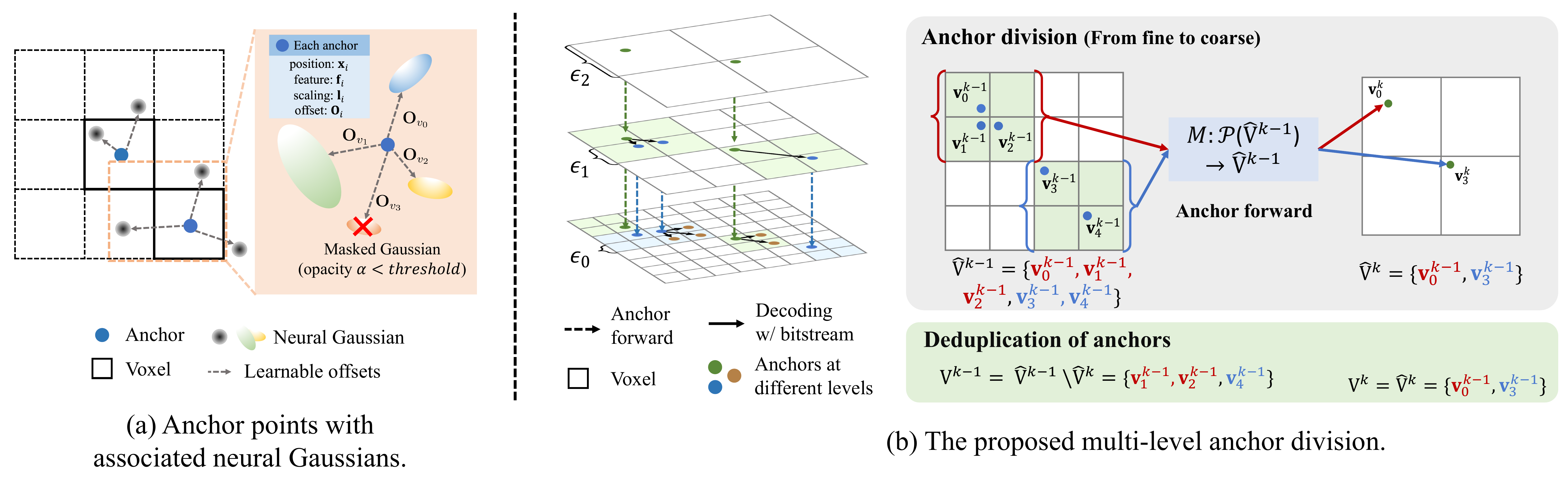}
    \caption{\textbf{(a)}: An illustration of the data structure we used following Scaffold-GS~\cite{scaffold}, where anchor points are used to extract common features of their associated neural Gaussians. \textbf{(b)}: The proposed multi-level division of anchor points. The decoded anchors from higher (coarser) levels are directly forwarded to the lower (finer) level to avoid duplicate storage. Besides, taking decompression as an example,  the already decoded anchors are used to predict anchors that are not decompressed yet, which greatly \wh{reduces} the spatial redundancy among adjacent anchors. (Best zoom in for details.)}
    \label{fig:anchor}
    \vspace{-0.2cm}
\end{figure}
\textbf{3DGS}~\cite{3DGS} utilizes a collection of anisotropic 3D neural Gaussians to depict the scene so that the scene can be efficiently rendered using a tile-based rasterization technique. Beginning from a set of Structure-from-Motion (SfM) points, each Gaussian point is represented as follows
\begin{equation}
    G(\x) = e^{-\frac{1}{2}(\x-\bm{\mu})^T\bm{\Sigma}^{-1}(\x-\bm{\mu})},
\end{equation}
where $\x$ is the coordinates in the 3D scene, $\bm{\mu}$ and $\bm{\Sigma}$ are the mean position and covariance matrix of the Gaussian point, respectively. To ensure the positive semi-definite of $\bm{\Sigma}$, $\bm{\Sigma}$ is represented as $\bm{\Sigma}=\mathbf{RSS}^T\mathbf{R}^T$, where $\mathbf{R}$ and $\mathbf{S}$ are scaling and rotation matrixes, respectively. Besides, each neural Gaussian has the attributes of opacity $\alpha \in \R^1$ and view-dependent color $\c \in \R^3$ modeled by spherical harmonic~\cite{zhang2022differentiable}. All the attributes, \ie, $[\bm{\mu}, \mathbf{R}, \mathbf{S}, \alpha, \c]$, in neural Gaussian points are learnable and optimized by the reconstruction loss of images rendered by the tile-based rasterization.

\textbf{Scaffold-GS~\cite{scaffold}.} While representing scenes with neural Gaussians greatly accelerates the rendering speed, the large amount of 3D Gaussians leads to significant storage overhead. To reduce the redundancy among adjunct 3D Gaussians, the most recent work, Scaffold-GS~\cite{scaffold}, proposes to introduce \textit{anchor points} to capture common attributes of local 3D Gaussians as shown in Fig.~\ref{fig:anchor} (a). Specifically, the \textit{anchor points} are initialized from neural Gaussians by voxelizing the 3D scenes. Each anchor point has a context feature $\f \in \R^{32}$, a location $\x \in \R^3$, a scaling factor $\l\in\R^3$ and $k$ learnable offset $\O \in \R^{k\times 3}$. Given a camera at $\x_c$, anchor points are used to predict the view-dependent neural Gaussians in their corresponding voxels as follows,
\begin{equation}
\{\c^i, \r^i, \s^i, \alpha^i\}_{i=0}^k = F(\f, \bm{\sigma}_{c}, \dvc)  
\end{equation}
where $\bm{\sigma}_{c} = ||\x-\x_c||_2$, $\dvc=\frac{\x-\x_c}{||\x-\x_c||_2}$, the superscript $i$ represents the index of neural Gaussian in the voxel, $\s^i, \c^i \in \R^3$ are the scaling and color respectively, and $\r^i \in \R^4$ is the quaternion for rotation. The positions of neural Gaussians are then calculated as
\begin{equation}
    \{\bm{\mu}^0, ..., \bm{\mu}^{k-1}\} = \x + \{\O^0, ..., \O^{k-1}\}\cdot \l,
\end{equation}

where $\x$ is the learnable positions of the anchor and $\l$ is the base scaling of its associated neural Gaussians. After decoding the properties of neural Gaussians from anchor points, other processes are the same as the 3DGS~\cite{3DGS}. By predicting the properties of neural Gaussians from the anchor features and saving the properties of anchor points only, Scaffold-GS~\cite{scaffold} greatly eliminates the redundancy among 3D neural Gaussians and decreases the storage demand.

\begin{figure}[t]
    \centering
    \vspace{-0.2cm}
    \includegraphics[width=1\linewidth]{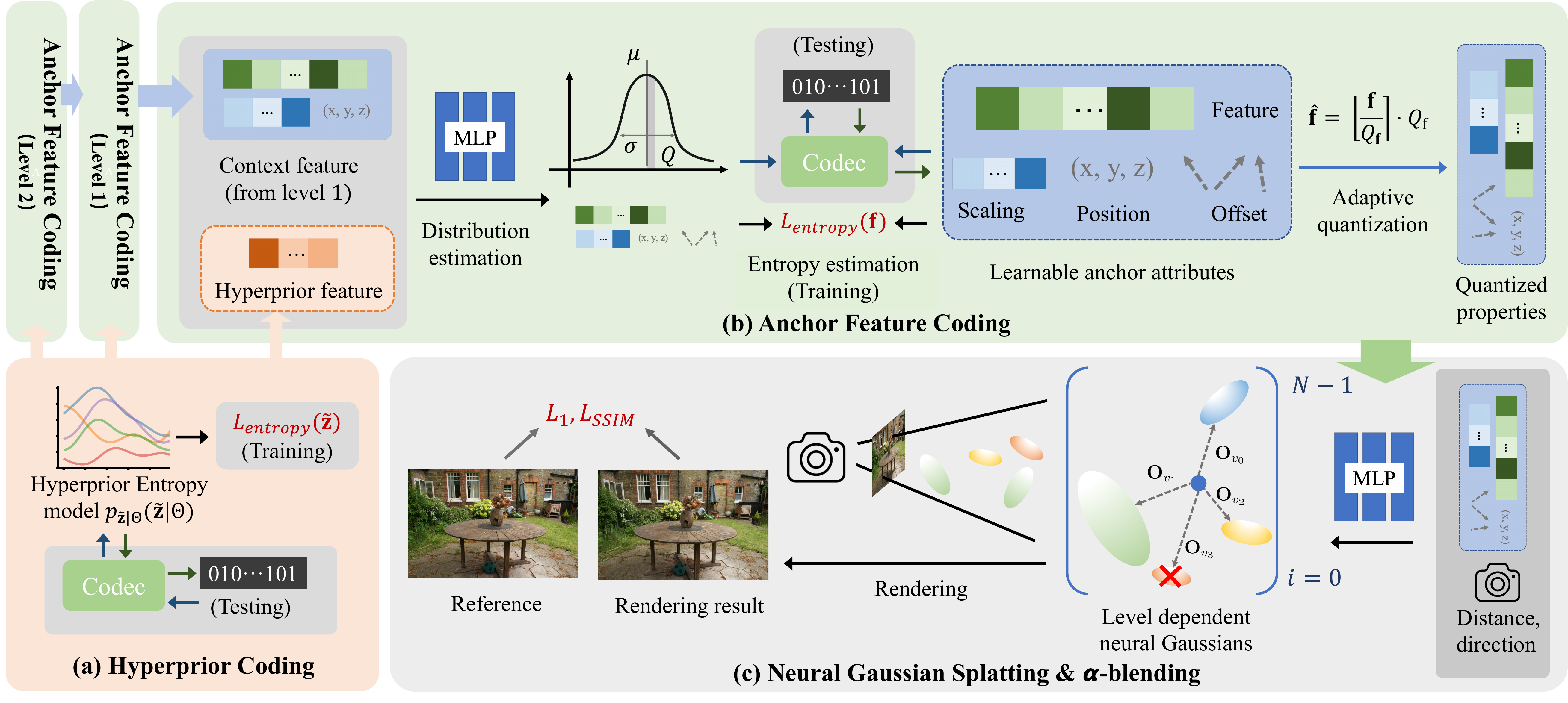}
    \caption{The overall framework of the proposed method includes three levels, \ie, $K=3$, to encode the anchors. The decoded anchors from a coarser level $i+1$ are used to encode the anchors in level $i$. Besides, hyperprior features are used to predict the properties of anchors at all levels. For training, after finishing the coding of all levels, the anchor features after adaptive quantization are used to predict properties of neural Gaussians. The rendering loss is calculated and optimized together with the entropy coding loss $\L_{entropy}$. For testing, after we decode anchor features from \wh{the} bit stream, the rendering is exactly the same with Scaffold-GS~\cite{scaffold} without introducing overhead.}
    \label{fig:framework}
    \vspace{-0.2cm}
\end{figure}

\section{Methodology}
The overall framework is shown in Fig.~\ref{fig:framework}. We first introduce how to divide anchors into levels with traceable mapping relationships among adjacent levels in Sec~\ref{sec:anchor_div}. Based on that, we present the entropy coding in an autoregressive way in Sec~\ref{sec:coding}, and the overall training objective in Sec~\ref{sec:training}.
\subsection{Anchor partitioning strategy}
\label{sec:anchor_div}

We attempt to divide $N$ anchors $\V=\{\v_i\}^N_{i=0}= \{(\x_{i}, \f_{i}, \l_{i}, \O_i)\}^N_{i=0}$ into $K$ disjoint levels, \ie, $\V=\V^0 \cup \V^1 \cup ... \cup \V^{K-1}$ and $\V^i \cap \V^j = \varnothing$, $\forall i \neq j$.
For each anchor set $\V^i$, it is expected to spawn over the whole scene, and be a relatively uniform downsampling of a finer set $\V^{i-1}$. 
Assume that we encode/decode the scene using the order from level $K-1$ to level $0$ where level $K-1$ is the coarsest level, we expect the mapping from $\V^{i}$ to $\V^{i-1}$ is traceable and easy to obtain. 
In other words, given an anchor $\v_i^k = (\x_{i}^k, \f_{i}^k,\l_{i}^k, \O_{i}^k)$ from a coarser level $k$, we expect to quickly locate $\v_i^k$'s adjacent anchor set $\{\v^{k-1}_{i}\}^{N^{k-1}_i}_{i=0}$ in level $k-1$, where $N^{k-1}_i$ is the number of adjacent anchors, \ie, the anchors in the same voxel of level $k-1$ with $\v_i^k$. As such, after decoding anchors at a coarser level, we can scatter the already decoded features to to-be-processed anchors as the prior for better coding efficiency.


To achieve the requirements above,  
we propose to utilize a simple yet effective way to divide anchors into levels using a ``bottom-up'' method. 
As shown in Fig.~\ref{fig:anchor} (b), given a set of anchors of the scene, we partition them into different fine-grained sets based on different voxel sizes $\epsilon_i$ as follows 
\begin{equation}
    \hat{\V}^k = \left\{M(\{\v_i^{k-1}: \hat{\x}_i^{k} = \hat{\x}\}): \hat{\x} \in  \left\{\hat{\x}^{k}_i: i=1,2,...,|\hat{\V}^{k-1}|\right\}  \right\},
\end{equation}
where $|\cdot|$ is the counting operation, $\hat{\x}^k_i$ is the anchor position after quantization using the voxel size of level $k$, and $M: \mathcal{P}(\hat{\V}^{k-1})\rightarrow \hat{\V}^{k-1}$ (where $\mathcal{P}$ is the power set) is a mapping function that selects a representative anchor to level $k-1$ from a set of anchors $\{\v_i^{k-1}: \hat{\x}_i^{k} = \hat{\x}\}$ that has the same position after quantization. The definition of $\hat{\x}_i^{k}$ and $M$ are as follows
\begin{equation}
\begin{split}
    \hat{\x}_i^{k} &= \left\lfloor \frac{\x^{k-1}_i}{\epsilon^{k}} \right\rceil \epsilon^{k}, \quad \x_i^{k-1} \in \hat{\V}^{k-1}\\
    M(\{\v_i^{k-1}: \hat{\x}_i^k &= \hat{\x}\}) = \v^{k-1}_{j} \text{ s.t. } j=\min\{i: \hat{\x}_i^k = \hat{\x}\},
\end{split}
\end{equation}
where $\hat{\V}^0$ is initialized as the whole anchor set $\V$. We select the anchor with the minimum index $\min\{i: \hat{\x}_i^k = \hat{\x}\}$ among a set of anchors in level $k-1$ that are in the same voxel. Besides, we filter out the repeated anchors among different levels as follows
\begin{equation}
    \V^2 = \hat{\V}^2, \V^1 = \hat{\V}^1 \backslash \hat{\V}^2, \V^0 = \hat{\V}^0 \backslash \hat{\V}^1
\end{equation}
where $\backslash$ is the set difference operation.
We keep the voxel size $\epsilon_0$ of the finest level (level $0$) the same as the initial value $\epsilon$, and set the voxel size of level $i$ to $\epsilon_i=\kappa_i \cdot \epsilon$. Since different scenes have different initial voxel sizes and anchor point distributions, using a fixed set of voxel scaling parameters $\{\kappa_i\}_{i=1}^K$ leads to a suboptimal performance. To avoid finetuning hyper-parameters for each scene, we propose to conduct a one-time parameter search after initializing anchors. Instead of directly setting the scale $\kappa_i$, we set a target ratio between level $i$ and $i+1$ and expect $\frac{|\V^{i+1}|}{|\V^i|} \approx \tau$. Since $|\V^i|$ decreases monotonically with $\kappa_i$, we can easily and efficiently determine the values of $\{\kappa_i\}_{i=1}^K$ using a binary search. We empirically find that the performance of models among different senses is relatively robust to the selection of $\tau$ (refer to Fig.~\ref{fig:abla_ratio}).

\subsection{Coding with entropy models}
\label{sec:coding}
After dividing the anchors into multi-levels, in this section, we discuss how we use the already decoded anchors to predict ones that are not decompressed yet and how to encode attributes of anchors to improve the coding efficiency. 

\textbf{Context modeling in anchor levels.}
To encode an anchor point $\v_i = (\x_{i}, \f_{i},\l_{i}, \O_i)$ into bitstreams efficiently using entropy coding, we need to estimate its distributions accurately. The core idea of the proposed method is to predict the properties of anchors additionally conditioned on already decompressed anchors. Taking the modeling of anchor feature $\f_i^k$ from the anchor $\v_i^k$ as an example, the details are as follows
\begin{equation}
\begin{split}
    p_{\f^k}(\f_{i}^k|\bm{\psi}_{i}^{k}) = (\mathcal{N} (\bm{\mu}_{i}^k,  \bm{\sigma}_{i}^k) * \mathcal{U}(-\frac{1}{2} \Delta_{i}^k,\frac{1}{2} \Delta_{i}^k))(\f_{i}^k), \quad
    \bm{\mu}_{i}^k, \bm{\sigma}_{i}^k, \Delta_{i}^k = F_{\f_i}^k (\bm{\psi}_i^k),
\end{split}
\end{equation}
where $F_{\f_i}^k$ is a MLP belonging to the level $k$, and $\bm{\psi}_i^k$ is the prior of the anchor $\v_i^k$ as follows
\begin{equation}
    \bm{\psi}_i^k = 
\left\{\begin{matrix}
[\x^k_{i}] & k = K-1 \\
[\f_{j}^{k+1}; \l_{j}^{k+1};\x^k_{i}], & k <K-1
\end{matrix}\right.,
\label{eq:context}
\end{equation}
where $[\cdot]$ is the concatenation operation among the channel dimension, $\f_{j}^{k+1}, \l_{j}^{k+1}$ are the feature and scaling from the adjacent anchor $\v_j^{k+1}$ in level $k+1$ that are already decoded as shown in Fig.~\ref{fig:anchor} (b). 

\textbf{Hyperprior feature for anchor.}  While introducing the position $\x_{i}^k$ contributes to \wh{predicting} the distribution of anchor features, it still lacks enough freedom to eliminate spatial redundancy. Therefore, we introduce a learnable hyperprior vector $\z_i \in \R^{50//h_c}$ for each anchor $\v_i$ where $h_c$ is a hyper-parameter to control the length of the hyperprior features. The hyperprior $\z_i$ is modeled using the non-parametric, fully factorized density model~\cite{balle2018variational} as follows:
\begin{equation}
   p_{\tilde{\z}|\Theta}(\tilde{\z}_i|\Theta) = \prod_{j=0}^{50//h_c-1}{ \left(p_{z_i^j|\Theta^{(j)}}(\Theta^{(j)})*\mathcal{U}\left(-\frac{1}{2}, \frac{1}{2}\right)\right)(\tilde{z}_i^j)},
\end{equation}
where $\tilde{\z}_i$ represents $\z_i$ with quantization noise, $j$ is the channel index and $\Theta$ is the network parameters for modeling the hyperprior. Since the hyperprior feature $\z_i$ is quantized into integers $\hat{\z}$ and jointly optimized to reduce the size of the bitstream, it only occupies a small portion of storage as shown in Table~\ref{tab:coding_anchor}. The final prior for coding features of the anchor $\v_i^k$  
 is $\hat{\bm{\psi}}_{i}^k = [\hat{\z}_i^k; \bm{\psi}_i^k]$.



\subsection{Training objective}
\label{sec:training}
The training objective of the proposed method is to jointly optimize the bitrate of coded anchor features and rendering loss measured by SSIM and L1 loss. The final training loss is 
\begin{equation}
\begin{split}
    \L = \L_{scaffold} +  \lambda_e \L_{entropy} + \lambda_m \L_m \\
\end{split}    
\end{equation}
where $\L_{scaffold}$ is the training loss of \cite{scaffold}, $\L_m$ is the masking loss from \cite{Joo} to regularize the masking loss of offsets of neural Gaussians $\O_v$, and $\L_{entropy}$ is the overall entropy loss that measures the cost of storing anchor properties defined as follows,
\begin{equation}
    \L_{entropy} = \mathbb{E} [-\log_2 p_{\tilde{\z}|\Theta}(\tilde{\z}_i|\Theta)] + \sum_{i=0}^{K-1}\left[-\log_2\left[\prod_{\f\in\{\f^k,\l^k, \O^k\}} p_{\f}(\f_{i}|\hat{\bm{\psi}}_i^k) \right]\right] 
\end{equation}
where the first term measures the cost of coding the hyperprior feature while the second term is the cost of coding features of anchor points in all the levels, and $\hat{\bm{\psi}}^k_i$ is the context feature that includes both the hyperprior feature $\z_i^k$ and the feature from already coded nearby anchor in level $k+1$ as illustrated in Eq.~\ref{eq:context}.



\section{Experiments}
\subsection{Implementation details}
We build our method based on Scaffold-GS~\cite{scaffold}. The number of levels is set to $3$ for all experiments and the target ratio among two adjacent iterations is $0.2$. $h_c$ is set to $4$, \ie, the dimension of the hyper-prior feature is a fourth of the anchor feature dimension. For a fair comparison, the dimension of the anchor feature $\f$ is set to $50$ following~\cite{chen2024hac} and we set the same $\lambda_m=5e-4$.
The setting of $\lambda_e$ is discussed in Appendix~\ref{sec:sup_detail} since different values are used to evaluate different rate-distortion tradeoffs. For a fair comparison, we use the same training iterations with Scaffold-GS~\cite{scaffold} and HAC~\cite{chen2024hac}, \ie, $30000$ iterations. Besides, we use the same hyperparameters for anchor growing as Scaffold-GS~\cite{scaffold} so that the final model has a similar or even smaller number of anchors, leading to faster rendering speed. More implementation details are in the supplementary materials. 

\subsection{Comparison with baselines}
\textbf{Baseline, metric, and benchmark.} We compare our method with 3DGS~\cite{3DGS}, Scaffold-GS~\cite{scaffold} and some representative 3DGS compression works, including Compact3DGS~\cite{Joo}, Compressed3D~\cite{Simon}, EAGLES~\cite{Sharath}, LightGaussian~\cite{zhiwen}, Morgenstern~\etal~\cite{Wieland}, Navaneet~\etal~\cite{KLNavaneet}, and HAC~\cite{chen2024hac}. The baseline methods include existing mainstream techniques, \eg, pruning~\cite{zhiwen, Joo}, codebooks~\cite{zhiwen, Joo, KLNavaneet, Simon}, and entropy coding~\cite{Sharath, chen2024hac, Joo, Wieland}, and includes the most recent works. We utilize PSNR, SSIM, and LPIPS~\cite{zhang2018perceptual} to evaluate the rendering qualities of different methods and report the storage size measured in MB. We evaluate the performance of the models on several real-scene datasets, including BungeeNeRF~\cite{BungeeNeRF}, DeepBlending~\cite{deepblending}, Mip-NeRF360~\cite{mip360}, and Tanks\&Temples~\cite{tant}. To more comprehensively evaluate the performance of our method, following the previous prototype~\cite{chen2024hac}, we use all $9$ scenes in Mip-NeRF360~\cite{mip360}. The detailed results of each scene are reported in the Appendix~\ref{sec:sup_detail}. To further evaluate the performance models among a wide range of compression ratios, we use Rate-Dsitoration (RD) curves as an additional metric.

\begin{table}[t] \scriptsize
    \centering
    \vspace{-0.2cm}
    \setlength\tabcolsep{1.45pt}
    \caption{The quantitative results obtained from the proposed method \textit{ContextGS} and other competitors. Baseline methods, namely 3DGS~\cite{3DGS} and Scaffold-GS~\cite{scaffold}, are included for reference. The intermediary approaches are specifically designed for 3DGS compression. Our methodology showcases two results representing varying size and fidelity tradeoffs, achieved through adjustment of $\lambda_e$. Highlighted in \colorbox{red!25}{red} and \colorbox{yellow!25}{yellow} cells are the best and second-best results, respectively. Size measurements are expressed in megabytes (MB).
    }
    \vspace{0.1cm}
    \scalebox{0.9}{
    \hspace{-0.2cm}
    \begin{tabular}{ll|cccc|cccc|cccc|cccc}
        \toprule
        \multicolumn{2}{l|}{\textbf{Datasets}} & \multicolumn{4}{c|}{\textbf{Mip-NeRF360~\cite{mip360}}} & \multicolumn{4}{c|}{\textbf{Tank\&Temples~\cite{tant}}} & \multicolumn{4}{c|}{\textbf{DeepBlending~\cite{deepblending}}} & \multicolumn{4}{c}{\textbf{BungeeNeRF~\cite{BungeeNeRF}}} \\
        \multicolumn{2}{l|}{\textbf{methods}} & psnr$\uparrow$    & ssim$\uparrow$   & lpips$\downarrow$ & size$\downarrow$   & psnr$\uparrow$   & ssim$\uparrow$   & lpips$\downarrow$ & size$\downarrow$   & psnr$\uparrow$   & ssim$\uparrow$   & lpips$\downarrow$ & size$\downarrow$ & psnr$\uparrow$    & ssim$\uparrow$   & lpips$\downarrow$ & size$\downarrow$   \\
        \bottomrule
        \multicolumn{2}{l|}{{3DGS~\cite{3DGS}\tiny{~(SIGGRAPH'23)}}}&27.49&\cellcolor{red!25}{0.813}&\cellcolor{red!25}{0.222}&744.7&23.69&0.844&{0.178}&431.0 & 29.42&0.899&\cellcolor{red!25}{0.247}&663.9 & 24.87&0.841&\cellcolor{red!25}{0.205}&1616    \\
        \multicolumn{2}{l|}{{Scaffold-GS~\cite{scaffold}\tiny{~(CVPR'24)}}}&27.50&0.806&0.252&253.9&23.96&\cellcolor{yellow!25}{0.853}&\cellcolor{yellow!25}{0.177}&86.50 & \cellcolor{yellow!25}{30.21}&{0.906}&0.254&66.00&{26.62}&{0.865}&0.241&183.0    \\  \hline
        \multicolumn{2}{l|}{{EAGLES}~\cite{Sharath}}&27.15&0.808&0.238&68.89&23.41&0.840&0.200&34.00 & 29.91&\cellcolor{red!25}{0.910}&\cellcolor{yellow!25}{0.250}&62.00&25.24&0.843&0.221&117.1    \\ 
        \multicolumn{2}{l|}{{LightGaussian~\cite{zhiwen}}}&27.00&0.799&0.249&44.54&22.83&0.822&0.242&22.43 & 27.01&0.872&0.308&33.94&24.52&0.825&0.255&87.28    \\  
        \multicolumn{2}{l|}{{Compact3DGS}~\cite{Joo}\tiny{~(CVPR'24)}}&27.08&0.798&0.247&48.80&23.32&0.831&0.201&39.43 & 29.79&0.901&0.258&43.21&23.36&0.788&0.251&82.60    \\  
        \multicolumn{2}{l|}{{Compressed3D~\cite{Simon}\tiny{~(CVPR'24)}}}&26.98&0.801&0.238&28.80&23.32&0.832&0.194&17.28 & 29.38&0.898&0.253&25.30&24.13&0.802&0.245&55.79  \\ 
        \multicolumn{2}{l|}{{Morgenstern~\etal~\cite{Wieland}} }&26.01&0.772&0.259&23.90&22.78&0.817&0.211&13.05 & 28.92&0.891&0.276&8.40&$-$&$-$&$-$&$-$    \\ 
        \multicolumn{2}{l|}{{Navaneet~\etal~\cite{KLNavaneet}}}&27.16&0.808&\cellcolor{yellow!25}{0.228}&50.30&23.47&0.840&0.188&27.97 & 29.75&0.903&\cellcolor{red!25}{0.247}&42.77&24.63&0.823&0.239&104.3   \\
        \multicolumn{2}{l|}{{HAC}~\cite{chen2024hac}}&{27.53}&0.807&0.238&\cellcolor{yellow!25}{15.26}&{24.04}&{0.846}&0.187&\cellcolor{yellow!25}{8.10} & 29.98&0.902&0.269&\cellcolor{yellow!25}{4.35}&26.48&0.845&0.250&\cellcolor{yellow!25}{18.49}    \\
        \hline
        \multicolumn{2}{l|}{Ours (low-rate)} & \cellcolor{yellow!25} 27.62 & 0.808 & 0.237 & \cellcolor{red!25}{12.68} & \cellcolor{yellow!25} 24.20 & 0.852 & 0.184 & \cellcolor{red!25} 7.05 & 30.11 & 0.907 & 0.265 & \cellcolor{red!25} 3.45  & \cellcolor{yellow!25} 26.90 & \cellcolor{yellow!25} 0.866 & \cellcolor{yellow!25} 0.222 & \cellcolor{red!25} 14.00  \\

        \multicolumn{2}{l|}{Ours (high-rate)} & \cellcolor{red!25} 27.75 &\cellcolor{yellow!25}{0.811} & 0.231 & 18.41 & \cellcolor{red!25} 24.29 & \cellcolor{red!25} 0.855 & \cellcolor{red!25} 0.176 & 11.80 & \cellcolor{red!25} 30.39 & \cellcolor{yellow!25} 0.909 & 0.258 & 6.60 & \cellcolor{red!25} 27.15 & \cellcolor{red!25} 0.875 & \cellcolor{red!25} 0.205 & 21.80  \\
        
    \bottomrule
    \end{tabular}}
    \label{tab:main_quantitative}
\end{table}

\begin{figure}[t]
    \centering
    \vspace{-0.2cm}
\includegraphics[width=\linewidth, trim=0 10 0 0, clip]{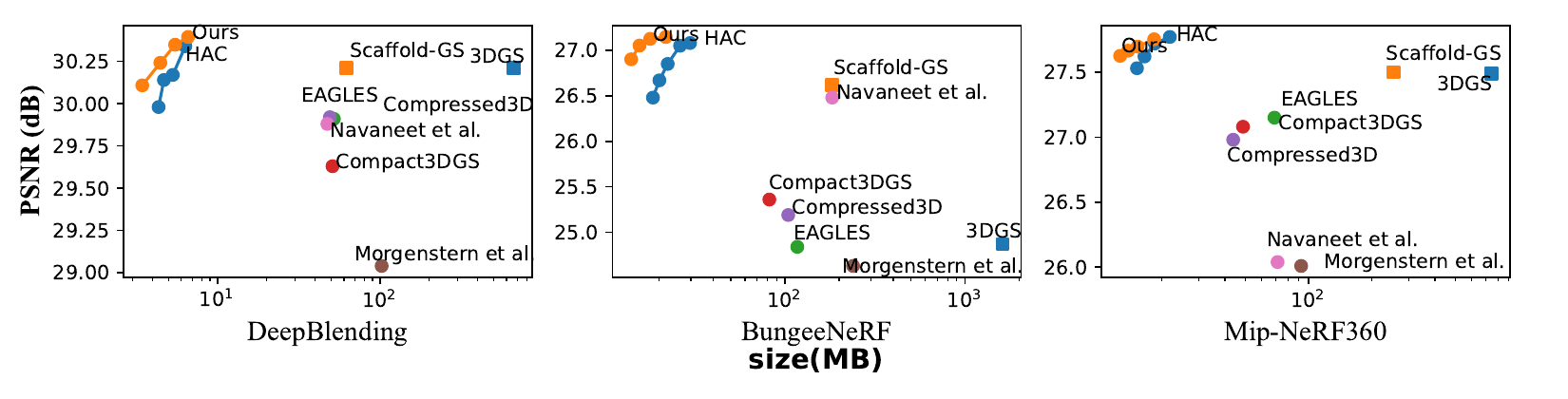}
\vspace{-0.5cm}
    \caption{The Rate-Distortion (RD) curves for quantitative comparison between our method with most recent SOTA competitors. It is worth noting that the x-axis is in $\log$ scale for better visualization.}
\vspace{-0.2cm}
    \label{fig:rd_curves}
\end{figure}

\begin{figure}[t]
    \centering
    \vspace{-0.2cm}
    \includegraphics[width=1\linewidth]{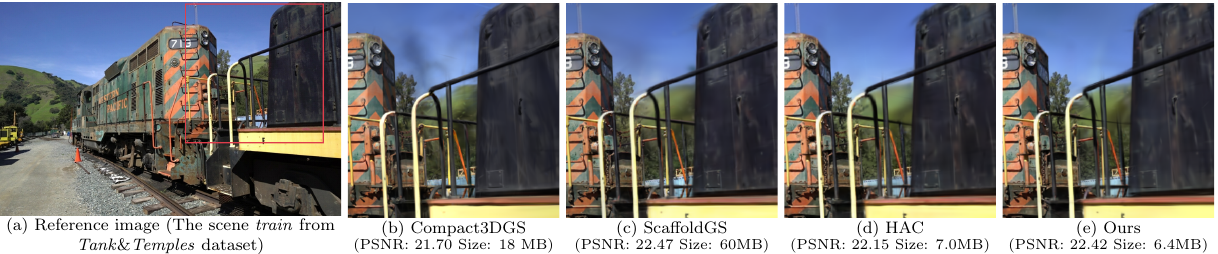}
    \includegraphics[width=1\linewidth]{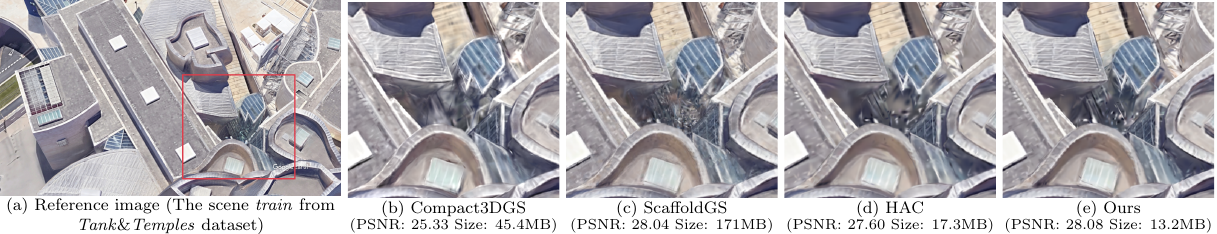}
    \caption{Visual comparisons between our method and baselines including Scaffold-GS~\cite{scaffold}, HAC~\cite{chen2024hac}, and Compact3DGS~\cite{Joo} on Bungeenerf~\cite{BungeeNeRF} and Tank\&Temples~\cite{tant}. We report the PSNR (dB) of the image and the size of the 3D scene. (Best zoom in for details.)}
    \label{fig:visual_comparsion}
    \vspace{-0.1cm}
\end{figure}

\textbf{Results.} As shown in Table~\ref{tab:main_quantitative}, the proposed method achieves significant improvement compared to our backbone method Scaffold-GS~\cite{scaffold} in terms of the size of the model, with a size reduction of $15\times$ in average. Besides, the proposed method also achieves higher storage efficiency compared to the most recent competitors for the 3DGS compression, \eg, HAC~\cite{chen2024hac}, Compressed3D~\cite{Simon}, and {Compact3DGS}~\cite{Joo}. It is worth noting that the proposed method also significantly improves rendering quality, even compared with the backbone model we use, \ie, Scaffold-GS~\cite{scaffold}. This further verifies the observation from previous works that appropriate constraints on neural Gaussians can contribute to the rendering quality, \eg, entropy constraints~\cite{chen2024hac} and pruning~\cite{yang2024spectrally}. Visual comparisons between the proposed method and other competitors are shown in Fig.~\ref{fig:visual_comparsion}. As shown in the figure, the proposed method achieves better rending quality with greatly reduced size compared with most recent 3D compression works and also the backbone model. Besides, a comparison of the RD curves among the proposed method and most recent competitors is shown in Fig.~\ref{fig:rd_curves}, where the proposed method achieves better performance in a wide range of compression ratios.

\begin{figure}[t]
    \centering
    \begin{minipage}[t]{0.39\textwidth}
        \centering
        \vspace{0pt}
\includegraphics[width=0.9\textwidth]{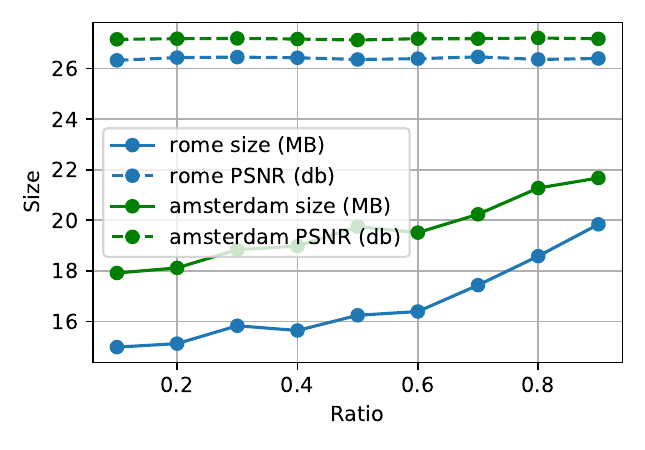}
        \vspace{-0.2cm}
        \caption{The ablation of different target ratio $\tau$ among different scenes. The PSNR remains relatively stable while the size of the scenes keeps increasing when increasing the $\tau$, which demonstrates the effectiveness.}
        \label{fig:abla_ratio}
    \end{minipage}
    \hspace{0.15cm}
    \begin{minipage}[t]{0.58\textwidth}
        \centering
        \captionof{table}{The ablation study of each component we proposed measured on BungeeNerf~\cite{BungeeNeRF} dataset. ``HP'' and ``CM'' represent the \textbf{h}yper\textbf{p}rior and anchor level \textbf{c}ontext \textbf{m}odel respectively. Ours w/o HP w/o CM can be roughly regarded as a Scaffold-GS~\cite{scaffold} model with entropy coding and masking loss~\cite{Joo}.}
        \scalebox{0.86}{
        \begin{tabular}{l|cccc}
        \toprule
         & Size (MB) & PSNR & SSIM & LPIPS\\
        \midrule
        Scaffold-GS~\cite{scaffold} & 183.0 & 26.62 & 0.865  & 0.241\\
        \midrule
        Ours w/o HP w/o CM & 18.67 & 26.93 & 0.867 & 0.222 \\
        Ours w/o CM & 15.03 & 26.91 & 0.866 & 0.223\\
        Ours w/o HP & 15.41	& 26.92	& 0.867 & 0.221\\
        Ours & 14.00 & 26.90 & 0.866 & 0.222 \\
        \bottomrule
        \end{tabular}}
        \label{tab:aba_component}
    \end{minipage}
\vspace{-0.3cm}
\end{figure}

\begin{figure}[t]
    \centering
    \vspace{-0.3cm}
    \begin{minipage}[t]{0.85\textwidth}
        \centering
        \captionof{table}{The ablation study of our method w/ and w/o reusing anchors from coarser levels, \ie, anchor forward technique, measured on BungeeNerf~\cite{BungeeNeRF} dataset.}
         \vspace{-0.1cm}
        \scalebox{0.85}{
        \begin{tabular}{l|cccc}
        \toprule
         & Size (MB) & PSNR & SSIM & LPIPS\\
        \midrule
        Ours w/o dividing into levels & 14.73 & 26.91 & 0.867 & 0.222 \\
        Ours w/o reusing anchors & 15.54 & 26.91 & 0.862 & 0.230 \\
        Ours & 13.80 & 26.89 & 0.867 & 0.222 \\
        \bottomrule
        \end{tabular}}
        \label{tab:aba_reuse}
    \end{minipage}
    \vspace{-0.2cm}
\end{figure}

\begin{figure}[t]
    \centering
    \hspace{0.15cm}
    \begin{minipage}[t]{1\textwidth}
        \centering
        \vspace{-0.3cm}
        \captionof{table}{The storage cost of each component and rending qualities of our method and baselines evaluated on the scene \textit{rome} in BungeeNeRF~\cite{BungeeNeRF} dataset. ``w/ APC'' represents using \textbf{a}nchor \textbf{p}osition \textbf{c}oding, \ie, using the hyperprior features to code the anchor positions. (The encoding/decoding time is measured on an RTX3090.)}
        \vspace{-0.1cm}
            \setlength{\tabcolsep}{0.3em}
        \scalebox{0.75}{
        \begin{tabular}{c|c|cccccccc|cc|cc}
        \toprule
         &\multirowcell{2}{Number of\\ Anchors (K)} & \multicolumn{8}{c|}{Storage Cost (MB)} & \multicolumn{2}{c|}{Fidelity} & \multicolumn{2}{c}{Speed (s)} \\
         \cmidrule(lr){3-14}
         & & Hyper & Position & Feature & Scaling & Offset & Mask & MLPs & Total & PSNR & SSIM & Encode & Decode \\
        \midrule 
Scaffold-GS~\cite{scaffold} & 61.9 & N/A & 7.08 & 75.16 & 14.18 & 70.88 & 2.362 & 0.047 & 186.7 & 26.25 & 0.872 & N/A & N/A \\
Ours (w/ APC) & 52.3 & 1.026 & \textbf{1.954} & 5.708 & 1.603 & 2.556 & 0.452 & 0.320 & \textbf{13.62} & 26.38 & 0.871 & 41.33 & 51.58 \\ 
Ours & 52.5 & 0.778 & 2.543 & 5.808 & 1.586 & 2.563 & 0.452 & 0.316 & 14.06 & 26.38 & 0.871 & \textbf{20.40} & \textbf{17.85} \\
        \bottomrule
        \end{tabular}}
        \label{tab:coding_anchor}
    \end{minipage}
    \vspace{-0.2cm}
\end{figure}

\subsection{Ablation studies and discussions}
\textbf{Evaluation of target ratio $\tau$ among adjacent levels.} To evaluate the performance of the proposed strategy that encodes anchors in a progressive way, we evaluate the performance of models trained under different ratios among adjacent levels. We disable the hyperprior feature to better explore the effect of different target ratios $\tau$. As shown in Fig.~\ref{fig:abla_ratio}, the PSNR remains relatively stable and the size gets relatively converged at the low ratio area. We select $\tau=0.2$ for all experiments.

\textbf{Ablation of each component.} We verify the effectiveness of two main components in our methods, \ie, anchor level context model and hyperprior features, and the results are shown in Table~\ref{tab:aba_component}. We build all the models on Scaffold-GS~\cite{scaffold} and set the model with the entropy constraint and the masking loss~\cite{Joo} as our baseline model, \ie, ``Ours w/o HP w/o CM''. It is worth noting that our baseline model significantly improves the storage efficiency compared with Scaffold-GS~\cite{scaffold} and even the latest SOTA methods. Both the proposed anchor-level context model and hyperprior feature for anchors significantly improve the compression rate compared with our strong baseline model, reducing the file size by 21\% and 10.17\%, respectively. Besides, using them together can further boost the performance with storage savings of 26.1\% and 92.5\% compared with the baseline we introduced above and Scaffold-GS~\cite{scaffold} respectively.

\textbf{Ablation of anchor forward.} A main difference between the proposed method and existing works for Level-of-Detail (LOD) techniques~\cite{ren2024octree} is that the proposed method can reuse the anchors of different levels, \ie, anchor forward in Fig.~\ref{fig:anchor} (b). For example, the anchors from different levels in \cite{ren2024octree} are stored separately. In contrast, the anchors from coarser levels are used in the final level (level $0$) in our method, \ie, in an autoregressive manner. To verify the effectiveness of the proposed method that reuses the anchors in coarser levels, we do an ablation study in Table~\ref{tab:aba_reuse}. As shown in the table, the model w/o reusing anchors of coarser levels to the finest level leads to serious redundancy, even slightly worse than the model w/o dividing anchors into different levels. This demonstrates the effectiveness of the proposed anchor forward technique for the anchor-level context model.

\textbf{Discussion on compressing anchor positions.} One can utilize the hyperprior feature $\z$ to predict the distribution of anchor positions and the anchor position can thereby be compressed using entropy coding. However, we find that the precision of the anchor position is essential to the performance of the model and an adaptive quantization strategy leads to serious performance degradation. While a fixed quantization width is feasible \wh{and} can retain the fidelity performance while effectively compressing the size of anchors, it leads to a greatly increased number of symbols that greatly decreases the coding speed. Since the anchor position only occupies a small portion of bitstreams as shown in Table~\ref{tab:coding_anchor}, we do not encode anchors into bitstreams in all the experiments.

\textbf{Analysis of inference and decoding time.}
The rendering speed after decompression is the same as or even faster than Scaffold-GS~\cite{scaffold} when the number of anchors is the same since we use the same data structure. However, as shown in Table~\ref{tab:coding_anchor}, we can achieve higher rendering quality with fewer anchors due to the use of masking loss~\cite{Joo} therefore achieving faster rendering speed. For the decoding time, while the proposed method involves autoregressive coding, which is usually very slow in image compression tasks~\cite{mentzer2018conditional} due to its serial characteristics, it adds neglectable overhead to our method in both training and decompression compared to other entropy-coding-based 3DGS compression methods, such as HAC~\cite{chen2024hac}. This is because, unlike autoregressive coding in image compression that predicts pixels/latent features one by one, introducing a loop of at least thousands of operations, we perform autoregressive coding group by group, introducing only a loop of 3 iterations. Additionally, there is no overlap of anchors among the coding of different levels, so the overall number of anchors to be processed remains the same as without dividing into levels. 

\section{Conclusion}
In this work, we introduce a pioneer study into utilizing the anchor-level context model in the compression of 3D Gaussian splatting models. We divide anchors into different levels and the anchors from coarser levels are first coded and then used to predict anchors that are not coded yet. Additionally, a hyperprior feature is used for each anchor that further reduces the channel-wised redundancy. 
Besides, we demonstrate that utilizing the proposed anchor forward technique, \ie, directly reusing the anchors from coarse levels to the final level, can achieve better performance than just using anchors of coarse levels as a prior.
Extensive experiments demonstrate that the proposed methods achieve better performance than SOTA and concurrent works.

{
\small
\bibliographystyle{abbrv}
\bibliography{neurips_2024.bib}
}

\clearpage
\newpage

\appendix

\section{Appendix / supplemental material}

\subsection{Broader impacts}
\label{sec:impact}
The social impacts of our work are mainly in three folds: 
\begin{enumerate}
    \item Accessibility: This work contributes to making advanced 3DGS~\cite{3DGS, scaffold} models more accessible to a wider audience. The reduced size of the 3DGS representation means that it can be more easily stored, transmitted, and processed on various devices, potentially democratizing access to high-quality rendering capabilities.
    \item Applications: Improved compression techniques for 3DGS could enhance various applications across industries, including virtual reality, gaming, medical imaging, and architectural visualization. These advancements may lead to more immersive experiences, better medical diagnostics, and more efficient design workflows.
    \item Open Access: The commitment to releasing the code fosters transparency and collaboration within the research community. Open access to the code allows other researchers and practitioners to build upon this work, accelerating innovation in the field of view synthesis and compression.
\end{enumerate}
We do not find serious negative impacts since our work is only for compression. There is no concerns regarding the misuse of our models or generating fake data.

\subsection{Limitations}
\label{sec:limitation}
A main and inevitable limitation of the proposed method is that the entropy coding process introduces extra computational costs to estimate the entropy of the anchor features during training encoding/decoding when saving/loading the 3D scene. For example, it requires extra time to decode the data of 3D scenes from the bitstream, making it challenging to start rendering at once when clicking the file. 

\subsection{More experimental details and results}
\label{sec:sup_detail}
We report the detailed comparison of our method w/ and w/o using the coding of anchor position in Table~\ref{tab:sup_anchor}.
To evaluate the performance among different rate-distortion (RD) tradeoffs, we utilize different $\lambda_{e}$, \ie, a larger $\lambda_{e}$ leads to a smaller model size. For a fair comparison, we use the same normalization of the hyper-parameter $\lambda_{e}$ as \cite{chen2024hac} by additionally utilizing the dimension of anchor features as the divisor so that the same $\lambda_{e}$ can lead to the similar bias towards the RD tradeoffs. The detailed results of each scene with different $\lambda_{e}$ are reported in Table~\ref{tab:sup_bung}, Table~\ref{tab:sup_tant}, Table~\ref{tab:sup_blending}, and Table~\ref{tab:sup_mip}.

\begin{table}[hb] \scriptsize
    \centering
    \setlength\tabcolsep{1.45pt}
    \caption{Quantitative results of the effect of anchor positon coding on our method. For our approach, we give two results of different size and fidelity tradeoffs by adjusting $\lambda_e$. 
    A smaller $\lambda_e$ results in a larger size but improved fidelity, and vice versa.}
    \scalebox{1.1}{
    \hspace{-0.2cm}
    \begin{tabular}{ll|cccc|cccc|cccc|cccc}
        \toprule
        \multicolumn{2}{l|}{\textbf{Datasets}} & \multicolumn{4}{c|}{\textbf{Mip-NeRF360~\cite{mip360}}} & \multicolumn{4}{c|}{\textbf{Tank\&Temples~\cite{tant}}} & \multicolumn{4}{c|}{\textbf{DeepBlending~\cite{deepblending}}} & \multicolumn{4}{c}{\textbf{BungeeNeRF~\cite{BungeeNeRF}}} \\
        \multicolumn{2}{l|}{\textbf{methods}} & psnr$\uparrow$    & ssim$\uparrow$   & lpips$\downarrow$ & size$\downarrow$   & psnr$\uparrow$   & ssim$\uparrow$   & lpips$\downarrow$ & size$\downarrow$   & psnr$\uparrow$   & ssim$\uparrow$   & lpips$\downarrow$ & size$\downarrow$ & psnr$\uparrow$    & ssim$\uparrow$   & lpips$\downarrow$ & size$\downarrow$   \\
        \bottomrule
        \hline
        \multicolumn{18}{c}{\textbf{w/ encoding anchors}} \\ 
        \multicolumn{2}{l|}{Ours (low-rate)} &  27.61 &  0.809 & 0.236 &  11.32 &  24.16 &  0.851 & 0.185 &  6.91 & 30.11 & 0.907 & 0.269 & 3.31 &  26.89 &  0.867 & 0.222 & 13.80 \\

        \multicolumn{2}{l|}{Ours (high-rate)} &  27.86 &  0.813 & 0.230 &  21.07 &  24.29 &  0.855 & 0.178 & 11.47 & 30.42 & 0.910 & 0.261 & 6.40 &  27.15 &  0.876 & 0.202 & 25.23 \\
        
        \midrule
        \multicolumn{18}{c}{\textbf{w/o encoding anchors}} \\ 

        \multicolumn{2}{l|}{Ours (low-rate)} &  27.62 & 0.808 & 0.237 & {12.68} &  24.20 & 0.852 & 0.184 &  7.05 & 30.11 & 0.907 & 0.265 &  3.45  &  26.90 &  0.866 &  0.222 &  14.00  \\

        \multicolumn{2}{l|}{Ours (high-rate)} &  27.75 &{0.811} & 0.231 & 18.41 &  24.29 &  0.855 &  0.176 & 11.80 &  30.39 &  0.909 & 0.258 & 6.60 &  27.15 &  0.875 &  0.205 & 21.80  \\
    \bottomrule
    \end{tabular}}
    \label{tab:sup_anchor}
\end{table}

\begin{table}[ht]
\centering
\caption{Per-scene results of our method on BungeeNerf~\cite{BungeeNeRF} dataset.}
\begin{tabular}{lccccc}
\toprule
     & Scene & Size & PSNR   & SSIM    & LPIPS    \\
\midrule
\multirowcell{7}{low-rate\\($\lambda=0.004)$}  & rome       & 13.9631 & 25.9751  & 0.8669   & 0.2143   \\
     & quebec     & 11.5909 & 30.1941  & 0.9337   & 0.1669   \\
     & pompidou   & 15.2258 & 25.4676  & 0.8469   & 0.2446   \\
     & hollywood  & 13.5183 & 24.5154  & 0.7772   & 0.3164   \\
     & bilbao     & 13.1537 & 28.0842  & 0.8877   & 0.1932   \\
     & amsterdam  & 16.5549 & 27.1694  & 0.8842   & 0.1956   \\
     &  \textbf{Average} & 14.0011 & 26.9010  & 0.8661   & 0.2218   \\
\midrule
\multirowcell{7}{high-rate\\($\lambda=0.001)$} & rome       & 21.7033 & 26.6297  & 0.8825   & 0.1929   \\
     & quebec     & 18.1517 & 30.5271  & 0.9400   & 0.1510   \\
     & pompidou   & 23.4748 & 25.6218  & 0.8546   & 0.2330   \\
     & hollywood  & 20.7082 & 24.7170  & 0.7876   & 0.3002   \\
     & bilbao     & 20.7686 & 27.9842  & 0.8928   & 0.1770   \\
     & amsterdam  & 26.0079 & 27.3979  & 0.8949   & 0.1754   \\
     & \textbf{Average} & 21.8024 & 27.1463  & 0.8754   & 0.2049   \\
\bottomrule
\end{tabular}
\label{tab:sup_bung}
\end{table}

\begin{table}[t]
\centering
\caption{Per-scene results of our method on DeepBlending~\cite{deepblending} dataset.}
\begin{tabular}{lccccc}
\toprule
     & Scene & Size & PSNR   & SSIM    & LPIPS   \\
\midrule
\multirowcell{3}{low-rate\\($\lambda=0.004)$} &  drjohnson	&	3.94 &	29.68 &	0.906 &	0.261  \\
&  playroom	&	2.96 &	30.53 &	0.907 &	0.269   \\
     & \textbf{Average} & 3.45 &	30.11 &	0.907 &	0.265  \\
\midrule
\multirowcell{3}{high-rate\\($\lambda=0.0005)$} &  drjohnson	&	7.80 &	29.86 &	0.909 &	0.249 \\
&  playroom	&	5.41 & 30.93 &0.910 & 0.268  \\
     & \textbf{Average} & 6.60 	&30.39 &	0.909 & 0.258 \\
\bottomrule
\end{tabular}
\label{tab:sup_blending}
\end{table}

\begin{table}[ht]
\centering
\caption{Per-scene results of our method on Tank\&Template~\cite{tant} dataset.}
\begin{tabular}{lccccc}
\toprule
     & Scene & Size & PSNR   & SSIM    & LPIPS    \\
\midrule
\multirowcell{3}{low-rate\\($\lambda=0.004)$}
 & train & 6.39 & 22.40 & 0.818 & 0.217 \\ 
 & truck & 7.72 & 26.00 & 0.885 & 0.150 \\
	& \textbf{Average} & 7.05 & 24.20 	& 0.852 &	0.184 \\
\midrule
\multirowcell{3}{high-rate\\($\lambda=0.0005)$}
 & train & 10.55 &22.53 &	0.823 &	0.208  \\ 
 & truck & 13.06 &	26.05 &	0.888 &	0.143  \\
	& \textbf{Average} & 11.80 &	24.29 & 	0.855 &	0.176   \\
\bottomrule
\end{tabular}
\label{tab:sup_tant}
\end{table}

\begin{table}[ht]
\centering
\caption{Per-scene results of our method on the Mip-NeRF360~\cite{mip360} dataset.}
\begin{tabular}{lccccc}
\toprule
     & Scene & Size & PSNR   & SSIM    & LPIPS    \\
\midrule
\multirowcell{10}{low-rate\\($\lambda=0.004)$}
 & bicycle  & 21.82 & 25.08 & 0.738 & 0.271 \\ 
 & bonsai   & 7.17  & 32.67 & 0.946 & 0.186 \\
 & counter  & 6.30  & 29.38 & 0.912 & 0.197 \\
 & flowers  & 16.71 & 21.31 & 0.576 & 0.377 \\
 & garden   & 18.78 & 27.32 & 0.845 & 0.148 \\
 & kitchen  & 7.00  & 31.28 & 0.925 & 0.131 \\
 & room     & 4.50  & 31.71 & 0.923 & 0.207 \\
 & stump    & 14.86 & 26.58 & 0.762 & 0.268 \\
 & treehill & 17.00 & 23.29 & 0.647 & 0.349 \\
 & \textbf{Average} & 12.68 & 27.62 & 0.808 & 0.237 \\
\midrule
\multirowcell{10}{high-rate\\($\lambda=0.0005)$}
 & bicycle  & 38.09 & 24.97 & 0.740 & 0.263 \\ 
 & bonsai   & 12.43 & 32.93 & 0.951 & 0.182 \\
 & counter  & 10.67 & 29.60 & 0.916 & 0.190 \\
 & flowers  & 28.27 & 21.18 & 0.571 & 0.378 \\
 & garden   & 31.26 & 27.39 & 0.851 & 0.136 \\
 & kitchen  & 12.44 & 31.69 & 0.931 & 0.123 \\
 & room     & 8.09  & 32.03 & 0.928 & 0.196 \\
 & stump    & 24.22 & 26.56 & 0.761 & 0.263 \\
 & treehill & 28.78 & 23.09 & 0.645 & 0.346 \\
 & \textbf{Average} & 21.58 & 27.72 & 0.811 & 0.231 \\
\bottomrule
\end{tabular}
\label{tab:sup_mip}
\end{table}

\end{document}